\documentclass[a4paper,conference]{IEEEtran}
\usepackage{siunitx}
\usepackage{amsmath}
\usepackage{amssymb}
\usepackage{gensymb}
\usepackage{booktabs}
\usepackage{multirow}
\usepackage{cite}
\usepackage{threeparttable}
\usepackage{nth}
\usepackage{verbatim}
\usepackage{pgffor}
\usepackage{pgfkeys}

\ifCLASSINFOpdf
   \usepackage[pdftex]{graphicx}
   \graphicspath{{/figures/}}
   \DeclareGraphicsExtensions{.pdf,.jpeg,.png,.jpg}
\else
\usepackage[dvips]{graphicx}
\fi

\usepackage{array}
\newcolumntype{H}{>{\setbox0=\hbox\bgroup}c<{\egroup}@{}}

\ifCLASSOPTIONcompsoc
 \usepackage[caption=false,font=normalsize,labelfont=sf,textfont=sf]{subfig}
\else
 \usepackage[caption=false,font=footnotesize]{subfig}
\fi

\usepackage{stfloats}
\fnbelowfloat

\usepackage{url}

\usepackage{xcolor}
\begin{document}
%
\title{How semantic and geometric information mutually reinforce each other in ToF object localization}



%
\IEEEoverridecommandlockouts
\author{\IEEEauthorblockN{Antoine Vanderschueren\IEEEauthorrefmark{1}\thanks{\IEEEauthorrefmark{1} Authors contributed equally to this work},
Victor Joos\IEEEauthorrefmark{1} and
Christophe De Vleeschouwer}
\IEEEauthorblockA{
ICTEAM Institute, 
UCLouvain,
 Louvain-la-Neuve, Belgium\\ Email: \{antoine.vanderschueren, victor.joos, christophe.devleeschouwer\}@uclouvain.be\\
 }}


\maketitle

\begin{abstract}
We propose a novel approach to localize a 3D object from the intensity and depth information images provided by a Time-of-Flight (ToF) sensor. Our method uses two CNNs. The first one uses raw depth and intensity images as input, to segment the floor pixels, from which the extrinsic parameters of the camera are estimated. The second CNN is in charge of segmenting the object-of-interest. As a main innovation, it exploits the calibration estimated from the prediction of the first CNN to represent the geometric depth information in a coordinate system that is attached to the ground, and is thus independent of the camera elevation. In practice, both the height of pixels with respect to the ground, and the orientation of normals to the point cloud are provided as input to the second CNN. Given the segmentation predicted by the second CNN, the object is localized based on point cloud alignment with a reference model.

Our experiments demonstrate that our proposed two-step approach improves segmentation and localization accuracy by a significant margin compared to a conventional CNN architecture, ignoring calibration and height maps, but also compared to PointNet++. 
\end{abstract}


%
\IEEEpeerreviewmaketitle

\section{Introduction}
Object localization is often a critical preliminary step in applications related to human behavior analysis.

In this context, the Time-of-Flight (ToF) camera offers the following non-negligible advantages : it provides a distance map, in addition to the reflected intensity, and, given its active nature, is relatively independent of lighting conditions. These advantages come with greatly reduced image resolution, and a shorter range. Overall, ToFs appear thus especially suited for the monitoring of small closed spaces like bedrooms and hospital rooms.

To the best of our knowledge, all methods for semantic segmentation and localization using ToF images use either only raw depth information \cite{Hazirbas,Jiang2018}, assume a pre-calibrated device and a fixed viewpoint \cite{Gupta2014}, or work directly on the point cloud resulting from the depth map \cite{Qi2016}.

As a main contribution, our work develops a solution that automatically calibrates itself with respect to the ground plane, so as to represent the scene independently of the camera elevation and distance to the scene. The proposed representation, turning depth information into height and local point cloud normal, is shown to improve the object segmentation accuracy, thereby leading to a better localization. 

\begin{figure*}[h!]
    \centering
    \includegraphics[width=\textwidth]{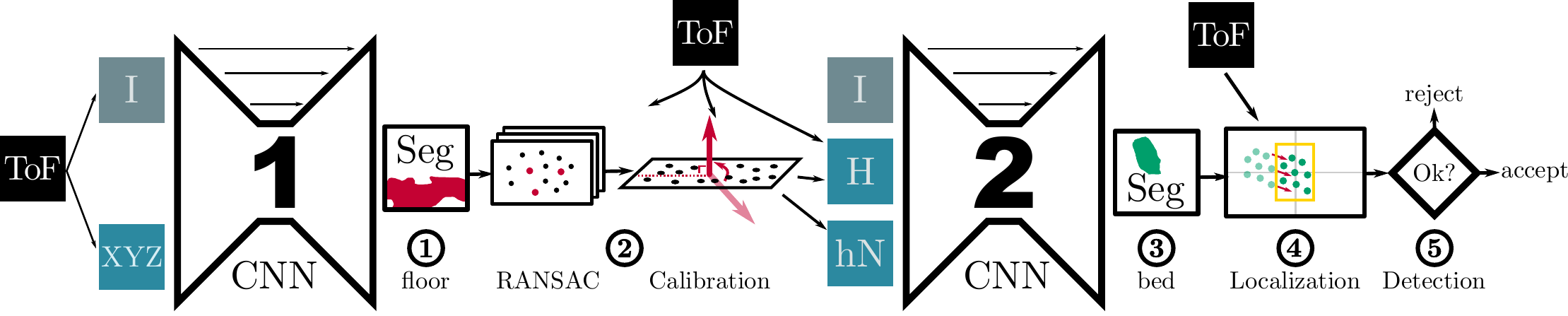}
    \caption{Overview of our method. The first segmentation CNN uses the intensity map (\texttt{I}) and spatial coordinates (\texttt{XYZ}) in the camera-centric viewpoint of the ToF. From the floor segmentation we derive the plane of the floor, and its normal. This enables us to use the height maps (\texttt{H}) and the normal maps (\texttt{hN}). The second CNN combines these maps with the intensity map to segment the object-of-interest. Segmented points are subsequently aligned with a reference model, in order to properly localize the object. The last step consists of a fitness detection, where cases of insufficient segmentation are refused.}
    \label{fig:overview}
\end{figure*}

Specifically, our calibration-aware approach follows the steps illustrated in \figurename{} \ref{fig:overview}. It first segments the floor based on a convolutional neural network (CNN) fed with raw depth and intensity maps. This initial segmentation allows us to estimate the ground plane equation in the coordinate system of the camera, so that the depth information can be transformed to height information with respect to the ground. While the previous system found its origin on the device, with the z axis pointing towards the scene, the ground-aligned system has the z axis parallel with the direction of gravity and is zero at ground-level.
The second step of our method consist in segmenting the object-of-interest while combining the same intensity with the newly computed height information and the estimated normal vectors in the ground-aligned system. At last, we use the segmentation of the object, and the new coordinate system, to localize the object in the scene.

Our method is validated on a \emph{practical} case using \emph{real} data. This case considers the localization of beds in nursing homes and hospital rooms. This dataset also showcases the need for a detection mechanism for correct or incorrect segmentation cases. We achieve this by computing how well the localization encompasses the segmentation output in order to detect erroneous localization.

Two main lessons are drawn from our experiments. First, convolutional networks surpass networks operating on point clouds like PointNet++ \cite{Qi2017a}. Second, our calibration-aware segmentation outperforms networks using only the raw camera output (depth and intensity).
By combining height, normal, and intensity maps we are able to gain 1.2\% Intersection-over-Union (IoU) on segmentation. This gain in segmentation accuracy enables a better localization, with a gain of a 1.4\% average precision.

This article is organized in 3 main sections. Section \ref{sec:sota} surveys the state-of-the-art. Section \ref{sec:method} then describes our method, and discusses its strengths and weaknesses compared to previous works. Section \ref{sec:results} validates our method on a real-life use case.

\section{State of the Art}\label{sec:sota}
The literature presents object localization using two distinct methodologies. Some recent efforts use 2D information in order to locate the object in 3D. Others are able to directly process a point cloud.

Our method belongs to the first category. Using 2D segmentation, we identify the pixels belonging to the object, and use the corresponding 3D points to compute the object position.

The rest of the section first considers methods relying on 2D segmentation. It then presents the approaches dealing with 3D, either based on convolutional or graph-based neural networks.

\subsection{2D Convolutional Segmentation}
ToF sensors are strongly related to RGB-D sensors. They mainly differ in their intensity signals: ToFs access a low-resolution reflected intensity signal, while RGB-D provides a high-resolution color image with more information density.  

The similarity between the signals, output by RGB-D and ToF, makes it relevant to extend the quite laconic SotA related to ToF segmentation (\cite{Holz2011,Maddalena2018}),  to the broader literature related to RGB-D segmentation \cite{Fooladgar2020}.

Common to all RGB-D signal segmentation methods is the importance of the fusion between the different information channels: color intensity and depth. \cite{Hazirbas} and \cite{Jiang2018} show the large impact of fusion on segmentation accuracy, the former with a simple encoder-decoder structure, the latter with residual blocks and skip-connections. Having noticed that a simple fusion by summation or concatenation wasn't enough to balance depth and color information correctly, \cite{Hu2019} uses a squeeze-and-excite \cite{Hu2018} attention module. We show in Section \ref{sec:results} that, in the case of ToF data, there is no benefit to fusion based on attention modules, which adds computational complexity without improving accuracy.

The disadvantage of the methods shown here is that they either only use depth information, and not more salient information like height, or make the hypothesis that an already gravity oriented point cloud is available. 

\subsection{3D Convolutional Localization}
3D Convolutional neural networks build on a voxelization of the point cloud. They suffer either from the lack of resolution induced by the use of big voxels, or from high sparsity in voxel information and high computational cost.

3D networks have first been considered for object localization in \cite{Milletari2016}. This pioneering work uses a U-Net \cite{Ronneberger2015} structure (like the one used in this paper), but substituting 2D for 3D convolutions.
Quite recently, \cite{Hou2020} has proposed to combine the high-resolution 2D color information, from RGB-D data, with a 3D neural network on point cloud data, leading to a precision gain of 2-3\%. However, we show in Section \ref{sec:ablation} that ToF intensity images contribute the \emph {least} of all the input types, to the final segmentation decision.

\cite{Song2016} has trained a network on synthetic data to complete the voxels that remain hidden when a single viewpoint is available. We have however observed that networks trained on synthetic data did not transfer well to real-life ToF data.

\subsection{Segmentation and Localization using Graph NNs}
To circumvent the excessive computational cost of 3D convolutions, graph neural networks work on connected points rather than on regular 2D or 3D matrices. Most implementations combine fully-connected layers and specialized pooling layers. Most point-based segmentation \cite{Liang2019,Li2019} and localization \cite{Landrieu2017,Qi2017,Qi2019,Shi2020} methods are based in part or in full on PointNet \cite{Qi2016} and PointNet++ \cite{Qi2017a}. Those approaches, although attractive in the way they represented the input point cloud, still show a lack of accuracy compared to convolutional methods on RGB-D and ToF images. We confirm this in Section \ref{sec:results} where PointNet++ is compared to our method.

\section{ToF$^2$-Net: Calibration before Localization}\label{sec:method}

To describe our method, we start by explaining the structure of our framework, and then delve deeper in its building blocks.

\subsection{ToF Data}

The ToF sensor gives us, for every pixel, information about reflected intensity and distance to the camera. Having access to the intrinsic parameters of the camera, we can express depth in terms of 3D position relative to the camera (\texttt{XYZ}). We can also estimate the normal vector to the point cloud surface in each point in this same coordinate system.

\subsection{Method Overview}
\begin{figure}
    \centering
    \subfloat[]{\includegraphics[width=0.48\columnwidth]{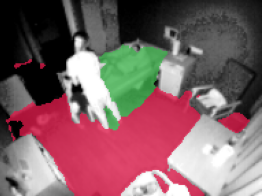}}
    \hfil
    \subfloat[]{\includegraphics[width=0.48\columnwidth]{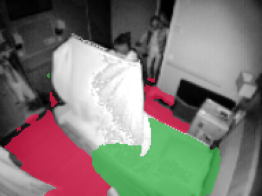}}
    
    \subfloat[]{\includegraphics[width=0.48\columnwidth]{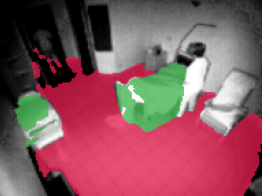}}
    \hfil
    \subfloat[]{\includegraphics[width=0.48\columnwidth]{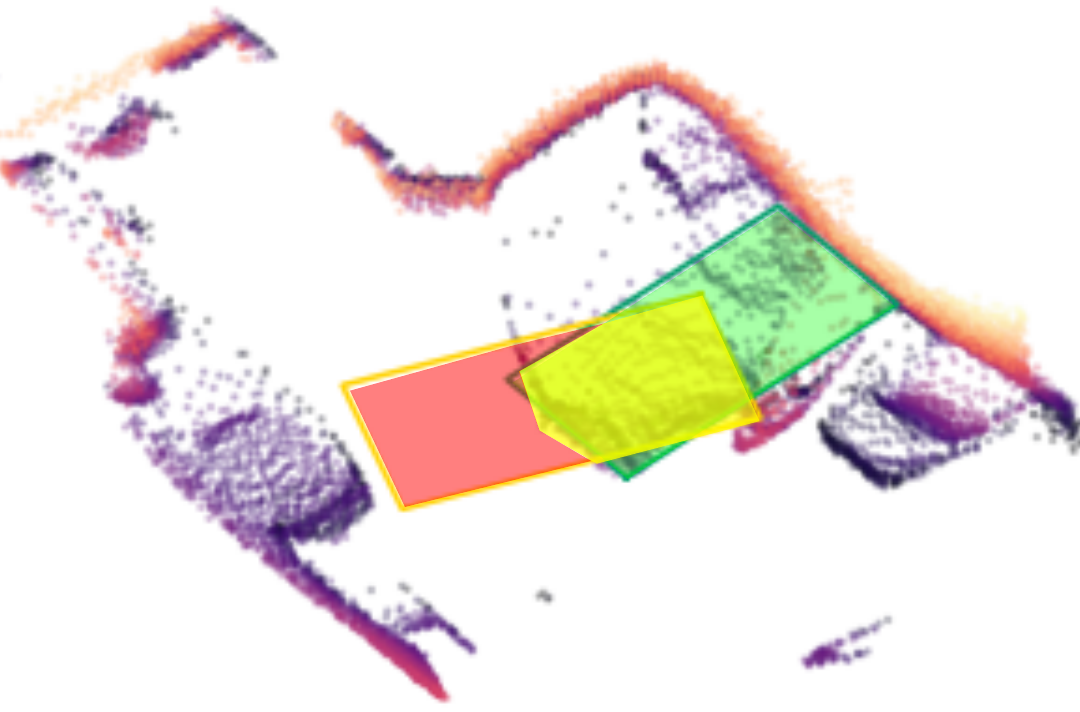}}
    \caption{(a-b) Examples of light (a) and heavy (b) occlusion. (c-d) Example of wrong segmentation (c). Estimated IoU will be low, as seen in the red region of (d) where there are no segmented points, and therefore the localization will be rejected. Note that rIoU$^b$ is not compared against the ground truth (green box) but against a projected rasterization of the object segmentation from (c). \emph{Best viewed in color.}}
    \label{fig:iou_estimation}
\end{figure}

Figure \ref{fig:overview} shows the five steps of our method:
\begin{enumerate}
    \item A first CNN uses the intensity and depth data to segment the floor pixels.
    \item The floor's plane equation in 3D space, relative to the camera, is obtained via Singular Value Decomposition (SVD) on the 3D coordinates of the floor pixels. The vector normal to the ground plane is defined by the smallest singular value. To make our algorithm more robust towards outliers, we embed SVD into a RANdom SAmpling Consensus (RANSAC) algorithm.
    \item A second CNN segments the object-of-interest.
    Since the ground plane equation is known, the height of every pixel can be computed from its distance to the camera, and the resulting pixel height map is fed to the network, together with the 3 components of the normal to the surface point cloud in each point.
    
    
    \item The points of the object are then fed into a localization algorithm that aligns them with a reference model.

    \item The quality of the matching between the object points and the model is used as confidence score for the localization, to detect cases of heavy occlusion or incorrect segmentation, as shown in \figurename{} \ref{fig:iou_estimation}.
\end{enumerate}

It should be noted that the calibration step (the \nth{2} step in the list above) allows us to move away from a camera-centric coordinate system towards a floor-centric point-of-view. To be more precise, starting from a coordinate system with z pointing away, x pointing down, and y perpendicular to the camera, we apply a rotation around the \emph{x}- and \emph{y}-axes in order to align the z-axis with the direction of gravity. Therefore, instead of expressing the data in a coordinate system fully dependent on the ToF placement, we express them in one where the coordinates are independent of the camera elevation. This gives us the two following benefits:
\begin{itemize}
    \item The object segmentation network learns the relationship between the geometric information and the object's mask more easily (as we'll show quantitatively and qualitatively in Section \ref{sec:results})
    \item The degrees of freedom during the localization step get reduced from 6 (3 rotations, 3 translations) to 3 (1 rotation, 2 translations).
\end{itemize}

In the remainder of this section, we'll detail how the CNNs used in steps 1 and 3 were constructed and trained, and implementation details for localization.
\subsection{Floor and Object Segmentation}\label{sec:method_seg}

In order to achieve a satisfying segmentation, we use a U-Net \cite{Ronneberger2015} shaped network. U-Net adapts an auto-encoder structure by adding skip-connections, which link feature maps of identical resolution from the encoder to the decoder. This allows the direct transfer of high resolution information by avoiding the lower resolution network bottleneck.

\begin{figure}
\centering
\includegraphics[width=0.9\columnwidth]{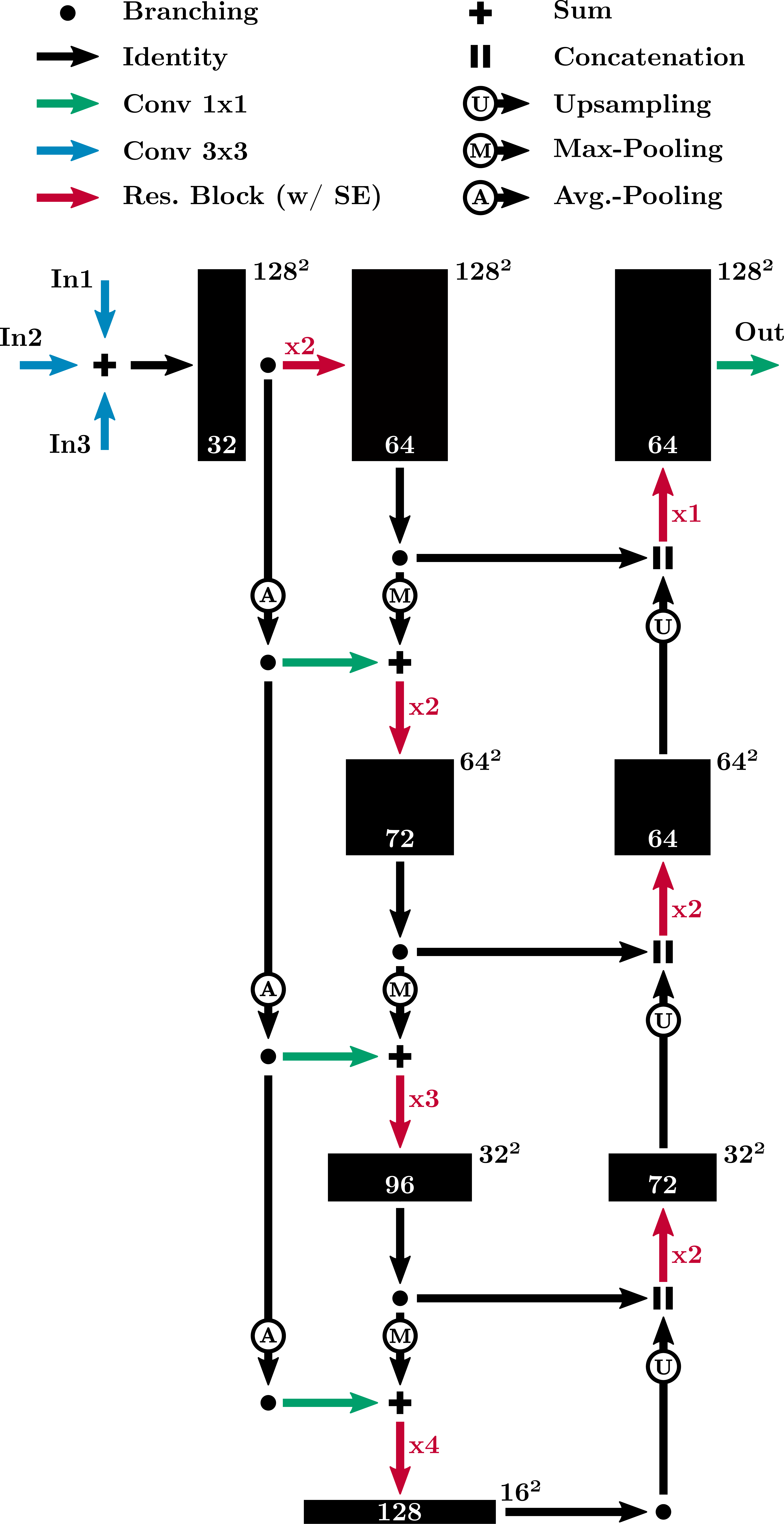}
\caption{ToF-Net Architecture. The boxes represent the feature maps, the number of channels are noted inside the box, the resolution outside. The architecture follows the U-Net model, with the addition of multi-resolution inputs, and block repetitions as shown in the figure. The different inputs are fused using a sum after a convolutional block.}
\label{fig:confunet}
\end{figure}

As illustrated in \figurename{} \ref{fig:confunet}, feature maps of identical resolution are said to be of the same \emph{level}.

For each level's structure, we settled on residual blocks \cite{He2015} followed by a \emph{Squeeze-and-Excite} module. The latter weighs every feature map individually before their sum. \cite{Hu2018} shows that this computationally inexpensive architectural modification brings significant performance gains.

We use different repetitions of each block at different levels with parameters from \cite{Yu2018, Jiang2018, chen2018}. The number of repetitions and feature maps at each level is detailed in \figurename{} \ref{fig:confunet}. 

Our networks are fed, as explained in the method overview, by a combination of the following input types: intensity (1D), normal (3D), depth (3D), and height (1D). To deal with different types of inputs, \cite{Hazirbas} proposed to fuse the inputs after they have been processed by a number of convolutions. This obviously increases the complexity. In order to evaluate whether the gain in performance associated to a late fusion is worth the complexity increase, our experimental section compares two different approaches:
\begin{itemize}
    \item Direct fusion (DF): every input type is passed once through a convolutional layer, such that they all possess the same number of feature maps. These feature maps are then summed once and fed to the rest of the network at every level, through down-scaling.
    \item Progressive fusion through attention modules (PFA) from \cite{Hu2019}: every input type has its own branch in the encoder. At the end of each level, those different branches are merged into a central trunk.
    This fusion module adds an attention mechanism by relying on the \emph{Squeeze-and-excite} (SE) module. Progressive fusion is almost 2 times slower than direct fusion.
\end{itemize}

Unless mentioned otherwise all results are for Direct Fusion networks since we'll show in Section \ref{sec:abfuse}, that PFA networks are heavily over-parameterized for our problem and ultimately lead to similar or worse performance.

\subsection{Localization and error estimation}\label{sec:method_loc}
As explained in the method overview, the points highlighted by our segmentation network are used to localize the object in a coordinate system where one axis is perpendicular to the floor, and the two remaining axes are respectively parallel and perpendicular to the intersection of the ToF image plane with the ground plane.

Since the bed has a simple shape we perform a very basic localization approach on a rasterized (discretized on a 5cm$^2$ resolution grid) 2D projection of the points of the bed segmentation on the ground plane.
We model the bed with a rectangle and initialize its center and rotation with the center of mass and principal direction, via SVD, of the projected points. Afterwards, we perform a local grid-search and select the model maximizing the intersection-over-union between the rasterized projected points (aggregating close points) and multiple 2D rectangular shapes (standard bed shapes). We thereby maximize estimated bounding-box IoU (rIoU$^b$).
By computing rIoU$^b$ on a rasterization grid, it is possible to detect cases of both under- and over-segmentation.



The rIoU$^b$ of the 2D rectangular model is then used as a confidence score to validate or reject the predicted bed localization (\nth{5} step). Finding the appropriate confidence threshold allows for deployment guarantees and enables human intervention for difficult cases. The higher this threshold is, the fewer cases of incorrect localization will be accepted, but the smaller the detection rate for correct localizations will be.

The localization is purposefully simple in order to explicitly evaluate our calibration-aware method, as well as providing this threshold-based system for classification of correct localization. As we'll see later in Section \ref{sec:results} it still seems sufficiently adequate for our case study.

\section{Results and Analysis}\label{sec:results}
This Section first introduces our validation methodology, including use case definition, training strategy, and quantitative metrics used for evaluation.  It then considers floor and object segmentation, respectively in Sections \ref{sec:res_floor} and \ref{sec:res_bed}. Eventually, the fusion of intensity and geometric inputs is further discussed in Section \ref{sec:ablation}.

\subsection{Validation methodology}
\subsubsection{Use case} 
Our method is evaluated on a ToF dataset, captured in nursing homes and hospital rooms. Our objective is to position the bed in the room, without any additional information but the intrinsic parameters of the camera and its output. To assess the performance of our system when it is faced with a new room layout or style, we divide our dataset in 7 subsets containing \emph{strictly} different institutions (hospitals or nursing homes). We apply cross-validation in order to systematically test the models on rooms that have not been used during training. 

Our database contains 3892 images of resolution 160x120. Those images come from 85 rooms belonging to 11 institutions. On average, 45 images with divers illumination, occlusions and (sometimes) bed positions are available per room.

In order to train and validate our models, we use manually annotated data, both for the device calibration and for the localization of the bed.\footnote{The tool developed for annotation is available at \url{https://github.com/ispgroupucl/tofLabelImg}.}

\subsubsection{Training}
During our cross-validation, we select 1 subset for testing, 1 subset for validation and 5 for training. We use data augmentation in the form of vertical flips and random zooming. The images are then cropped and rescaled to fit a 128x128 network input resolution.

Normal vectors are estimated from the 10 closest neighbors that are within a 10 cm radius  of each point. They are expressed in a 3D coordinate space either relative to the camera (\texttt{N}) or to the floor (\texttt{hN}).

All our segmentation networks result from a hyper-parameter search evaluated on each validation set. The average performance of all 7 subsets obtained on the test set is then presented. The networks are trained using AdamW \cite{Loshchilov2019} by choosing the learning rate and the weight decay in $\left\{\num{1e-3}, \num{5e-4}, \num{1e-4} \right\}$ and $\left\{\num{1e-4}, \num{1e-5}, \num{1e-6} \right\}$ respectively. The learning rate is divided by 10 at epochs 23 and 40, while training lasts for 80 epochs, with a fixed batch size of 32. We implement our segmentation models using PyTorch \cite{Paszke2019} and have published our code at \url{https://github.com/ispgroupucl/tof2net}. For the PointNet++ segmentation model, we use the model described in \cite{Qi2017a} and implemented by \cite{Wijmans2018}. We never start from a pre-trained network as pre-training shows poor performance on our dataset in all cases.

Finally in order to avoid initialization biases, the values presented in \emph{every} table are always the average of 5 different runs.

\subsubsection{Metrics}
Distinct metrics are used to assess segmentation, calibration and bed localization.

Segmentation predictions are evaluated using Intersection-over-Union (IoU), which is defined as
\begin{align}
\text{IoU} &= \frac{|\texttt{GT} \cap \texttt{PRED}|}{|\texttt{GT} \cup \texttt{PRED}|}\label{eq:iou}\\
&= \frac{\texttt{TP}}{\texttt{TP}+\texttt{FP}+\texttt{FN}},
\end{align}

where \texttt{GT} and \texttt{PRED} correspond to the manual annotation and the model prediction, respectively. The recall $\left(\frac{\texttt{TP}}{\texttt{TP+FN}}\right)$ and precision $\left(\frac{\texttt{TP}}{\texttt{TP+FP}}\right)$ are also considered separately, to better understand the nature of segmentation failures.


We evaluate the extrinsic camera calibration using absolute angles between the ground normal predicted by our model and the ground truth.

Object localization is also evaluated using IoU, but instead of comparing sets of pixels, we compare the 2D bounding-box obtained after localization to the ground truth, projected on a common plane, using IoU (denoted IoU$^b$). The projection on a common plane is necessary to compensate for slight differences due to possible calibration errors i.e. different ground plane equations.

In practice, IoU$^b$ that lie below 70\% correspond to localization not sufficiently accurate to support automatic behavior analysis, typically to detect when a patient leaves the bed. Hence we consider 70\% as a relevant localization quality threshold, and evaluate our methods based on the Average Precision at this threshold (AP@.7). In addition, the Area Under Curve (AUC@.7) measures the correct localization predictions (true positives) as a function of the incorrect ones (false positives) when scanning the confidence score given by the estimated rIoU$^b$, explained in Section \ref{sec:method_loc}.


\subsection{Floor segmentation and calibration accuracy}\label{sec:res_floor}

\begin{table}
\centering
\begin{threeparttable}
    \caption{Floor Segmentation and Calibration Results.}
    \label{tab:floor}
    \begin{tabular}{@{}>{\bfseries}l>{\ttfamily}lH*{2}{l}@{}}
        \toprule
        \normalfont{Method} & \normalfont{Inputs\tnote{1}} & Fusion\tnote{2} & IoU$_{floor}$ (\%) & Angle Diff. (\degree{})\\\midrule
        RANSAC & XYZ & - & - & 13.2\\
        PointNet++ \cite{Qi2017a} & XYZ+I & - & 83.6 & \hphantom{1}1.2 \\
        ToF-Net & I & - & 75.6 & \hphantom{1}3.3 \\
        ToF-Net & I+XYZ & DF & \textbf{87.2} & \hphantom{1}1.3 \\
        ToF-Net & I+XYZ+N & DF & 86.9 & \hphantom{1}\textbf{1.0} \\
        \bottomrule
    \end{tabular}
    \begin{tablenotes}
    \item Best results in \textbf{bold}
    \item [1] \texttt{I} denotes intensity, \texttt{XYZ} spatial coordinates, and \texttt{N} estimated normal.
    \end{tablenotes}
\end{threeparttable}
\end{table}

Table \ref{tab:floor} compares the different floor segmentation models in terms of segmentation and ground normal accuracy. It also presents the ground normal error obtained when applying RANSAC directly on the whole set of points (since a majority of points are floor points, one might expect RANSAC will discard outliers and estimate the ground plane equation without segmentation). We observed that the global RANSAC performed reasonably on some rooms, with an error smaller than 10\degree{} on 45\% of the samples. However it failed to find the floor on 55\% of the samples, which leads to a mean error of 13\degree{}.

Looking at the segmentation IoUs, we see that PointNet++ has better accuracy than a convolutional network that only uses the intensity (\texttt{I}) as input signal. However, adding the \texttt{XYZ} point coordinates  (as defined in the coordinate system of the camera) as input to the convolutional model is enough to surpass PointNet++. The geometric information provided by the normals, \texttt{N}, does not improve the segmentation IoU.

In terms of ground normal estimation, our proposed model leads to an absolute error of less than 1.3\degree{}, which is precise enough for our use case.
We also note that even though the segmentation maps made by PointNet++ and ToF-Net-\texttt{I+XYZ+N} are significantly worse than the ones predicted by ToF-Net-\texttt{I+XYZ}, the final floor angle error is similar for all networks with access to geometric information.

\subsection{Bed segmentation and localization}\label{sec:res_bed}
\begin{table}
\centering
\begin{threeparttable}
    \caption{Bed Segmentation and Localization Results}
    \label{tab:bed}
    \begin{tabular}{@{}>{\bfseries}l>{\ttfamily}lH llll@{}}\toprule

        \normalfont{Method} & \normalfont{Inputs\tnote{1}} & & IoU(\%) & IoU$^b$(\%) & AP@.7 & AUC@.7 \\\midrule
         PointNet++\cite{Qi2017a} & XYZ+I & -
                                    & 44.6 & 60.8 & 46.2 & 43.5\\
         ToF-Net & I          & -   & 65.3 & 67.1 & 60.0 & 58.6 \\
         ToF-Net & I+XYZ      & DF  & 69.9 & 75.4 & 76.2 & 74.7 \\
         ToF-Net & I+XYZ+N    & DF  & 70.4 & \underline{77.1} & \underline{79.2} & \underline{77.7}\\

         \addlinespace
         ToF$^2$-Net & I+H    & DF  & \underline{71.0} & 76.1 & 78.5 & 77.1\\
         ToF$^2$-Net & I+H+hN & DF  & \textbf{72.1} & \textbf{77.2} & \textbf{80.6} & \textbf{79.1}\\
         \bottomrule
    \end{tabular}
    \begin{tablenotes}
    \item Best results in \textbf{bold}, \nth{2} best \underline{underlined}
    \item [1] \texttt{I} denotes intensity, \texttt{XYZ} spatial coordinates, \texttt{H} height, and  \texttt{N} and \texttt{hN} normal maps relative to different viewpoints.
    \end{tablenotes}
\end{threeparttable}
\end{table}

\begin{figure*}
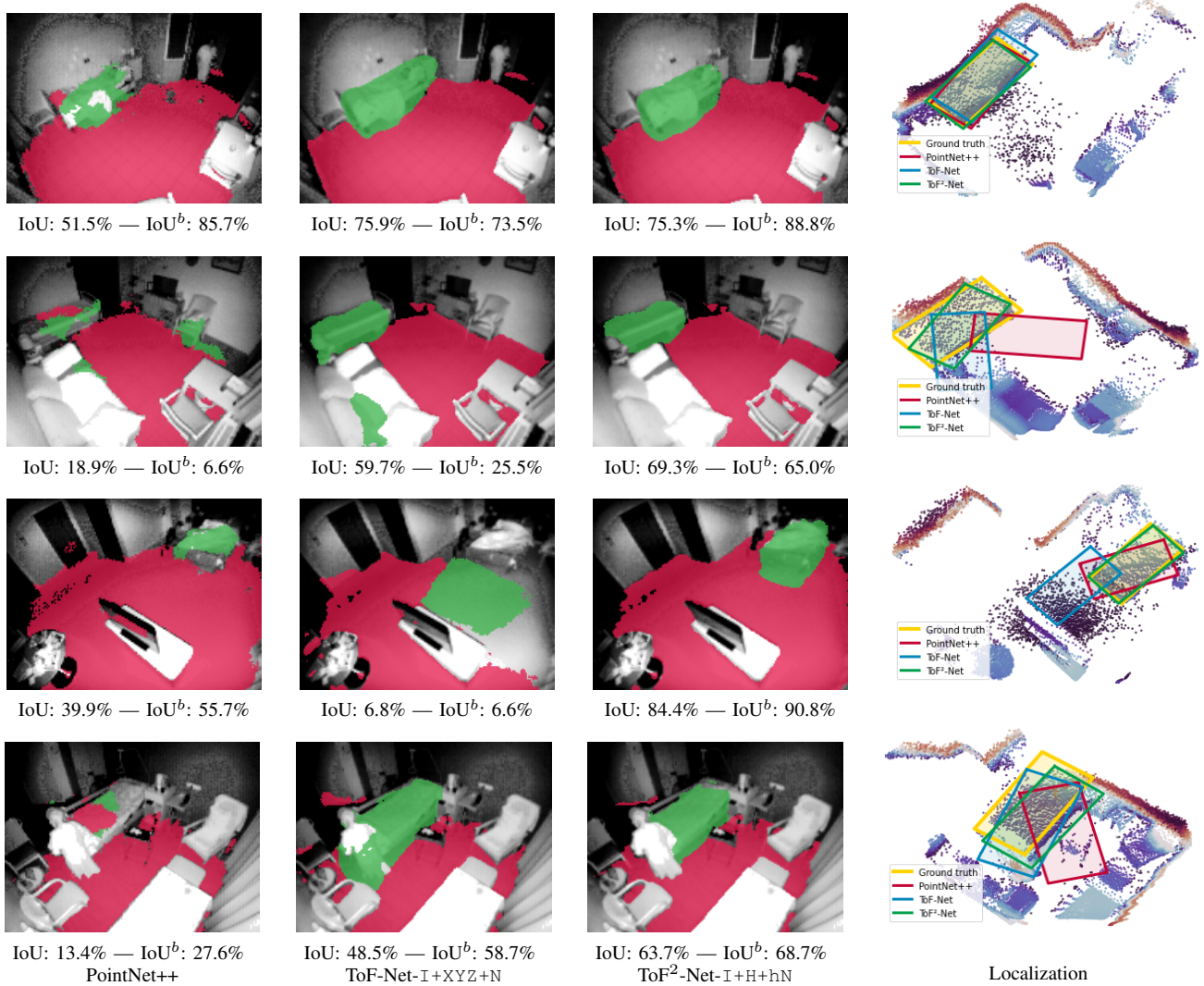

    \centering
    \captionsetup[subfigure]{labelformat=empty,farskip=0pt}
    \input{figures/comparison/values}
    \def\methods{{2020-07-07T093853_v4_pointnetxyz}/{PointNet++},{2020-07-11T214213_v4_XYZ_N_mIn5}/{ToF-Net-\texttt{I+XYZ+N}},{2020-07-10T185203_v4_H_hN_mIn5}/{ToF$^2$-Net-\texttt{I+H+hN}}}
    \def\images{{Fold1--ACIS-StJoseph-room_A.0.04_42},{Fold2--Ardennes-room_103_38},{Fold3--ORPEA-Bonaparte-room_201_0},{Fold6--SLG-LesRecollets-room_1.1.021_28}}
    \foreach \image [count=\nimages] in \images {%
    \foreach \method/\name [count=\ni] in \methods {%
        \ifnum\ni>1
            \hfil
        \fi
        \ifnum\nimages<4
            \subfloat[][\centering IoU: \pgfkeysvalueof{/iou/\method/\image}\% | IoU$^b$: \pgfkeysvalueof{/ioub/\method/\image}\%]{\includegraphics[width=0.2\textwidth]{figures/comparison/\image_\method.png}}
        \else
            \subfloat[][\centering IoU: \pgfkeysvalueof{/iou/\method/\image}\% | IoU$^b$: \pgfkeysvalueof{/ioub/\method/\image}\%
            \newline
            \name
            ]{%
            \includegraphics[width=0.2\textwidth]{figures/comparison/\image_\method.png}}%
        \fi%
        \ifnum\ni=3%
            \hfil
            \ifnum\nimages<4
            \subfloat{\includegraphics[width=0.25\textwidth]{figures/comparison/loc_\image_\method.png}}
            \else
            \subfloat[][\centering\newline Localization]{\includegraphics[width=0.25\textwidth]{figures/comparison/loc_\image_\method.png}}
            \fi
        \fi
    }
    }
    
    \caption{Qualitative results of segmentation and localization. The first three columns show the segmentation results of PointNet++, the best ToF-Net, and the best ToF$^2$-Net. The last column shows the localization results for the same models in a top view with height encoded in point color. Floor points have been removed for easier visualization. PointNet++ has a tendency to correctly locate the bed in most cases, but is unable to fully segment the bed. ToF-Net shows correct segmentation in most cases, but has problems locating beds in complex locations, for example in the third row. ToF-Net also suffers from oversegmentation, as shown in the second and last rows. ToF$^2$-Net is able to segment most cases correctly, but can suffer from biased segmentation for the localization, as in row 2 and 4. The last column shows clearly that a correct segmentation leads to good localization. \emph{Best viewed in color.}}
    \label{fig:quali}
\end{figure*}

Table \ref{tab:bed} summarizes the results of bed segmentation and localization for different modalities and network inputs. We compare single-step segmentation models (PointNet++, ToF-Net), which predict bed localization in one inference pass, to our calibration-aware proposal (ToF$^2$-Net) described in Section \ref{sec:method}.

\subsubsection{Segmentation}
The object localization relies heavily on the object segmentation accuracy. For this reason we first compare the different methods in terms of segmentation in the third column of Table \ref{tab:bed}.

We were initially surprised by the very low accuracy of PointNet++, even compared to the intensity-only baseline. However, this can be explained by the fact that the shape of the bed is more complex than the planar floor geometry. In addition, ToF are known to induce large measurement disparities, making it harder for a geometry-based neural network to correctly learn the object's geometry.

Looking at our calibration-aware method, we see that it systematically outperforms 1-step methods. The calibration-aware approach is especially accurate when the normal vectors are defined in the ground-centric coordinate system. Although normal vectors do help in the 1-step setting, they add a lot of noise when \emph{not} height-encoded and hinder accuracy.

\figurename{} \ref{fig:quali} shows qualitative results between the different best-performing segmentation networks. The results show the advantage of our calibration-aware approach compared to the one-step method. PointNet++ has the particularity of being very good at localization but having a hard time providing a full segmentation. This highlights the disparity in its results for segmentation compared to localization.

\subsubsection{Localization}
Table \ref{tab:bed} also displays the metrics for object localization. When looking at the bounding-box IoU$^b$ or the average precision at a threshold of 0.7 IoU (AP@.7), we observe the same trends as with the segmentation IoU. The gap between a formulation with and without floor calibration information is however reduced to 0.1\% for IoU$^b$ but widened to 1.4\% for AP@.7. 

Finally we look at our detection metric which determines if we accept or reject a bed localization. As explained in Section \ref{sec:method} we look at the AUC@.7 to determine whether we'll easily be able to identify erroneous samples. The maximal possible value in case of an always perfect detection would correspond to the AP@.7. The drop-off going from AP@.7 to AUC@.7 is approximately 1.5\% across the board. This relatively small decline shows the accuracy of our confidence threshold rIoU$^b$. Our calibration-aware approach still comes out on top with a 1.4\% lead.

\subsection{Analysis of Fusion and Network Inputs}\label{sec:ablation}
\subsubsection{Progressive Fusion through Attention Modules}\label{sec:abfuse}
\begin{table}
\centering
\begin{threeparttable}
    \caption{Comparison of IoU, bounding-box IoU and computational complexity for DF vs PFA on ToF$^2$-Net}
    \label{tab:fusion}
    \begin{tabular}{>{\bfseries}l>{\ttfamily}l lll}
        \toprule
        \multirow{1}{*}{\normalfont{Fusion\tnote{1}}} & 
        \multirow{1}{*}{\normalfont{Inputs\tnote{2}}}
         & IoU(\%) & IoU$^b$(\%) & Complexity\\\midrule

         Baseline & I & 65.3 & 67.1 & 1.00x \\\addlinespace
         DF & I+XYZ & \textbf{69.9} & \textbf{75.4} & \textbf{1.02x}\\
         DF & I+H+hN & 72.1 & \textbf{77.2} & \textbf{1.02x} \\
         \addlinespace
         PFA & I+XYZ & 69.1 & 72.5 & 1.84x\\
         PFA & I+H+hN & \textbf{72.2} & 76.8 & 2.31x\\
         \bottomrule
    \end{tabular}
    \begin{tablenotes}
    \item Best results for each input type are in \textbf{bold}
    \item [1] Fusion is either direct (DF) or through progressive fusion attention modules (PFA).
    \item [2] \texttt{I} denotes intensity, \texttt{XYZ} spatial coordinates, \texttt{H} height, and  \texttt{N} and \texttt{hN} normal maps relative to different viewpoints.
    \end{tablenotes}

\end{threeparttable}
\end{table}
Table \ref{tab:fusion} compares the Direct and Progressive fusion approaches described in Section \ref{sec:method_seg} with the complexity reflecting the relative increase in number of floating point operations. We note that the PFA-models use more resources and don't show any significant performance uplift. On the contrary, they are almost always detrimental.

\subsubsection{Input Ablation}
\begin{table}
\centering
\begin{threeparttable}
    \caption{Ablation of the inputs of ToF$^2$-Net trained on \normalfont{\texttt{I+XYZ+H+N+hN}}}
    \label{tab:ablation}
    \begin{tabular}{lllllll}
        \toprule
        & \multirow{2}[2]{*}{\texttt{All}}   & \multicolumn{5}{c}{\textbf{Independent removal of}} \\\cmidrule{3-7}
        &  & \texttt{I} & \texttt{N} & \texttt{XYZ}  & \texttt{hN} & \texttt{H}\\\midrule
        \textbf{IoU}         & 70.6 & 67.1 & 66.0 & 63.4 & 57.3 & 27.6\\
         \bottomrule
    \end{tabular}
    \begin{tablenotes}
    
    \item \texttt{I} denotes intensity, \texttt{XYZ} spatial coordinates, \texttt{H} height, and  \texttt{N} and \texttt{hN} normal maps relative to different viewpoints.
    \end{tablenotes}

\end{threeparttable}
\end{table}
In order to strengthen the intuition that spatial information, and the way it is presented is important, we'll now look at a network trained with every possible input type: \texttt{I+XYZ+N+H+hN}. In that way, the network gets the opportunity to select its preferred representation of the spatial information.

The Direct Fusion, explained in more details in Section \ref{sec:method}, is the summation of every input convolved once, $\left(\sum C^i_{in}\right)$. 

Each $C^i_{in}$, can thus individually be zeroed-out. Doing this outright leads to activation-range scaling issues since the network isn't retrained. Thus every channel in the removed $C^i_{in}$ is replaced by the mean of its activations. We do this for every input-type independently and present the results in Table \ref{tab:ablation}.

Presented with different encodings for the spatial information, the network consistently prefers height-encoded information. This confirms our hypothesis that height-encoded spatial information is a more digestible form of spatial information.

\section{Conclusion}
We propose ToF$^2$-Net, a calibration-aware method that successfully localizes beds in hospital and nursing home rooms. Our method estimates the extrinsic parameters of the device in order to have access to height maps and height-encoded normal vectors. We extensively show that this way of encoding ToF data leads to better performing CNNs and more digestible spatial information. Along the way we also demonstrate that fusion-modules are not necessary for this low-resolution mixed-input-type use-case and showcase the shortcomings of PointNet++. 




We recommend future work on input analysis using both visualization and network-based input-importance learning.
\IEEEtriggeratref{22}


\bibliographystyle{IEEEtran}
\bibliography{IEEEabrv,references.bib}

\end{document}